%% file: plan2scene_cvpr21.tex
\documentclass[final]{cvpr}
\input{macros}
\begin{document}
\title{\task: Converting Floorplans to 3D Scenes}
\author{Madhawa Vidanapathirana \quad Qirui Wu \quad Yasutaka Furukawa \quad Angel X. Chang \quad Manolis Savva\\
Simon Fraser University
}

\maketitle

\input{sec/abstract}
\input{sec/intro}
\input{sec/related}
\input{sec/task}
\input{sec/data}
\input{sec/method}
\input{sec/experiments}
\input{sec/conclusion}

\input{sec/acks}

{\small
\bibliographystyle{plainnat}
\setlength{\bibsep}{0pt}
\bibliography{egbib}
}


\end{document}

%% file: macros.tex
\usepackage{times}
\usepackage{epsfig}
\usepackage{graphicx}
\usepackage{amsmath}
\usepackage{amssymb}
\usepackage{booktabs}
\usepackage{makecell}
\usepackage{bbding} 
\usepackage{subcaption}
\usepackage{paralist}
\usepackage[ruled]{algorithm2e} 

\usepackage{adjustbox}
\usepackage[table,xcdraw]{xcolor}
\usepackage[numbers,sort,compress]{natbib}
\usepackage{comment}
\usepackage[pagebackref=true,breaklinks=true,colorlinks,bookmarks=false]{hyperref}

\usepackage{cleveref}

\def\cvprPaperID{0000} 
\def\confYear{CVPR 2021}

\newcommand{\topic}[1]{\noindent \textbf{#1}}
\newcommand{\mypara}[1]{\vspace{2pt}\noindent\textbf{#1}}
\newcommand{\ra}[1]{\renewcommand{\arraystretch}{#1}}

\newcommand{\task}[0]{Plan2Scene\xspace}

\newcommand{\Crop}{\texttt{Crop}\xspace}
\newcommand{\Retrieve}{\texttt{Retrieve}\xspace}
\newcommand{\NaiveSynth}{\texttt{NaiveSynth}\xspace}
\newcommand{\Synth}{\texttt{Synth}\xspace}

\newcommand{\Color}{\textsc{Color}\xspace}
\newcommand{\Freq}{\textsc{Freq}\xspace}
\newcommand{\Subs}{\textsc{Subs}\xspace}
\newcommand{\FID}{\textsc{FID}\xspace}
\newcommand{\Tile}{\textsc{Tile}\xspace}

\newcommand{\denselist}{\itemsep 0pt\parsep=0pt\partopsep 0pt\vspace{-\topsep}}

%% file: sec/abstract.tex
\begin{abstract}
We address the task of converting a floorplan and a set of associated photos of a residence into a textured 3D mesh model, a task which we call \emph{Plan2Scene}.
Our system 1) lifts a floorplan image to a 3D mesh model; 2) synthesizes surface textures based on the input photos; and 3) infers textures for unobserved surfaces using a graph neural network architecture.
To train and evaluate our system we create indoor surface texture datasets, and augment a dataset of floorplans and photos from prior work with rectified surface crops and additional annotations.
Our approach handles the challenge of producing tileable textures for dominant surfaces such as floors, walls, and ceilings from a sparse set of unaligned photos that only partially cover the residence.
Qualitative and quantitative evaluations show that our system produces realistic 3D interior models, outperforming baseline approaches on a suite of texture quality metrics and as measured by a holistic user study.
\end{abstract}

%% file: sec/intro.tex
\section{Introduction}

Digital 3D scene representations of interiors are key to emerging application areas such as AI assistants, online product marketing, and augmented reality.
Private residences are predominantly designed with CAD software.
However, texture-mapped 3D scene models of the built interiors are rarely available.
Despite recent progress in indoor reconstruction techniques and depth-sensing hardware, the state-of-the-art in room layout inference and photogrammetry-based 3D modeling is still
far from reliable and practical for non-expert users.

\begin{figure}[t]
    \includegraphics[width=\linewidth]{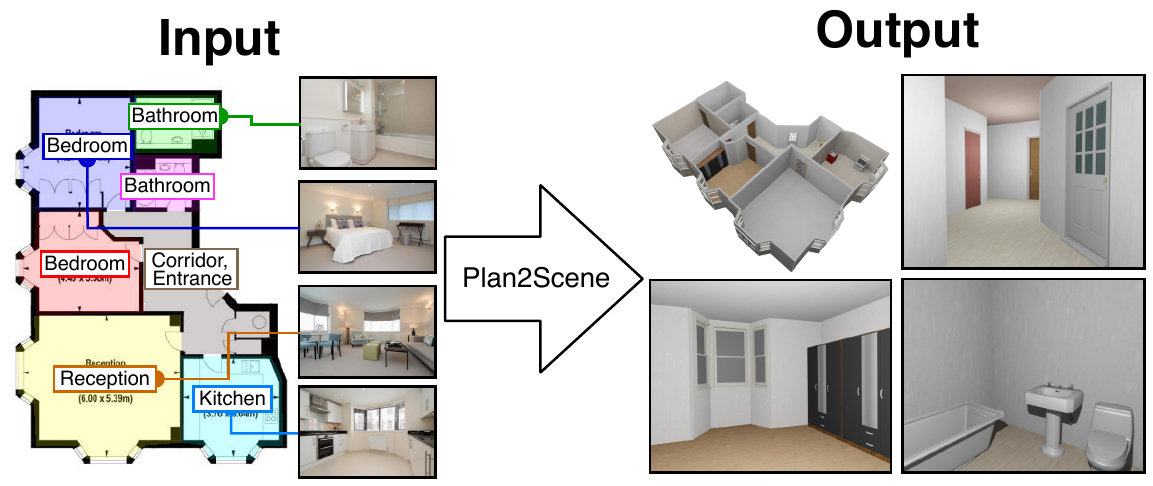}
    \caption{Our system addresses the \task task by converting a floorplan and set of photos to a textured 3D mesh.}
    \label{fig:small-teaser}
\end{figure}

This paper explores a novel path for 3D interior digitization by utilizing a residential floorplan and a sparse set of photos without camera pose as input, which are prevalent in online real estate listings.
More precisely, we present the \emph{\task} task: conversion of a residential floorplan and photos of a residence to a textured 3D scene model (see \Cref{fig:small-teaser}).
In the context of this task, we focus on texturing of architectural surfaces.
This task is challenging due to 1) the lack of camera poses for the photos;
2) the challenge of photometric calibration under varying lighting conditions
and 3) limited photo coverage, leaving many surfaces unobserved.
The \task task allows us to articulate these challenges for residential interiors and to identify suitable texture appropriateness metrics.

Our key idea is to model the architectural surfaces and identify appropriate textures for each surface.
We treat photos as sparse and partial observations of surface textures and formulate a texture inference task, instead of relying on exact camera poses and  
texture-mapping as in prior work~\cite{ApartmentsCVPR15}.
Textures for observed surfaces are generated using an encoder-decoder architecture.
Unobserved surfaces (i.e., surfaces for which a photo is not available) are handled by a graph neural network (GNN) that propagates information while learning inter/intra-room consistency.

The paper also makes three dataset contributions.
First, we extend an existing database of floorplans and photos (Rent3D)~\cite{ApartmentsCVPR15}
by annotating more room boundaries, assigning more photos to the rooms, and annotating object-icons indicated on the floorplans.
We also curate two datasets of textures of common indoor substances (e.g., `painted walls', `tiles', `carpets') from various online sources for training our texture synthesis method.

Through qualitative and quantitative evaluations using a suite of metrics characterizing texture appropriateness and quality, and a user study, we demonstrate that the proposed approach outperforms baselines including rectified image patch texturing, and direct texture retrieval.
We release all our code, data, and pretrained models to the community.\footnote{\url{https://3dlg-hcvc.github.io/plan2scene/}}

%% file: sec/related.tex
\section{Related work}

\mypara{Room layout estimation and 3D reconstruction.}
A long line of work exists on coarse 3D layout estimation from indoor perspective images or panoramas: \citet{zhang2014panocontext,zou2018layoutnet,yang2019dula,sun2019horizonnet,zhang2020geolayout}.
In contrast to our task, these methods 1) generate coarse geometry for a room as a set of planes 2) process only a single image and a single room instead of an entire residence; and 3) simply project image pixels onto the layout planes without separating objects from architectural surfaces.
There is also a rich literature of house-scale 3D reconstruction and modeling methods.
Different methods take as input a set of RGB images~\cite{furukawa2009reconstructing,cabral2014piecewise}, RGBD videos~\cite{huang20173dlite,zollhofer2018state}, RGBD panoramas~\cite{mura2014automatic}, dense point clouds~\cite{xiao2014reconstructing} or partial reconstructions~\cite{lin2019floorplan}.
In contrast, our input is a floorplan and a sparse set of RGB images without precise camera pose partially covering the interior, which is the typical data available on real estate websites and we produce as output a 3D textured mesh for the entire residence.

\mypara{Texture synthesis.}
There is a rich literature of work on exemplar-based texture synthesis.
We refer readers to the survey by \citet{akl2018survey}.
Recent work has adapted neural networks architectures for this task, allowing the use of embeddings to represent textures~\cite{Li2017DiversifiedTS,Oechsle2019TextureFL}.
\citet{Chen2020IntelligentH3} generate tileable textures from text descriptions.
Other work has focused on inferring SVBRDF models from a single image~\cite{li2017modeling,li2018materials} or from multi-illumination images~\cite{murmann2019dataset}.
We adopt an approach inspired by recent work that utilizes a compact texture embedding~\cite{henzler2020neuraltexture} as we rely on embedding propagation to generate textures for unobserved surfaces.

\begin{figure}
    \centering
    \subfloat[\centering Rent3D]{\includegraphics[width=3.6cm]{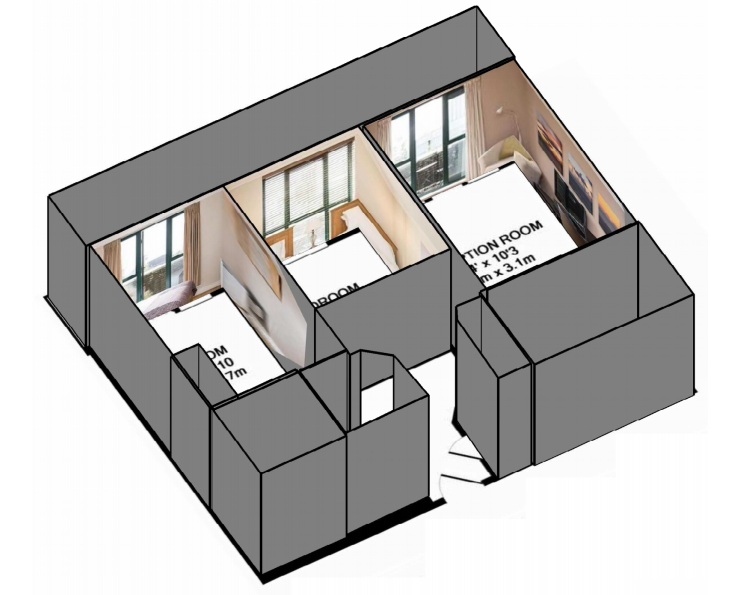}}%
    \qquad
    \subfloat[\centering Plan2Scene]{\includegraphics[width=3.6cm]{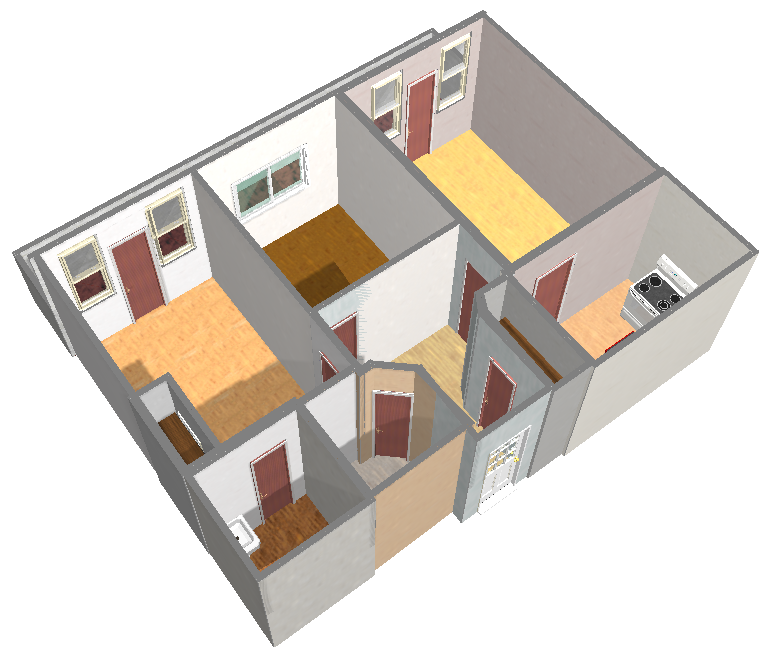}}%
    \caption{Comparison of (a) output from \citet{ApartmentsCVPR15} to (b) our output. We produce textured 3D meshes of the residence that do not exhibit distortions due to direct warping of photos onto walls, and that cover all surfaces.}
    \label{fig:related:rent3d-vs-ours}
\end{figure}

\mypara{3D scene generation from photo and floorplan data.}
Our \task task is closely related to the work of \citet{izadinia2017im2cad} on IM2CAD and \citet{ApartmentsCVPR15} on Rent3D.
IM2CAD infers a 3D room layout as well as 3D object placements from a single image.
Objects and walls are colored using the medoid of each color channel in the input RGB image, making it impossible to represent common material types such as wood, tile, or carpet.
Furthermore, the approach only handles surfaces and objects visible in a single input image.
In contrast, our approach generates textured materials for both observed and unobserved surfaces.
We also handle multi-room interiors.
Rent3D has a similar problem formulation as us, taking a floorplan and set of photos as input and producing a coarse 3D mesh.
However, it focuses on estimating the camera pose for each image and directly projecting pixels onto the mesh as an appearance model.
This results in unrealistic rooms with sofas, beds and other objects projected onto surfaces (see \Cref{fig:related:rent3d-vs-ours}).

%% file: sec/task.tex
\section{\task Task}

\begin{figure*}
    \centering
    \includegraphics[width=\linewidth]{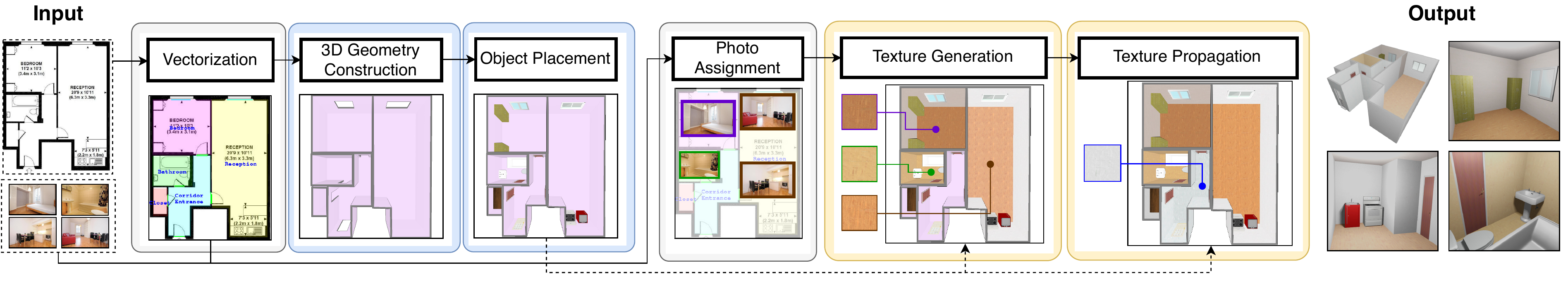}
    \caption{In the \task task we produce a textured 3D mesh of a residence from a floorplan and set of photos. This process involves several steps: floorplan vectorization, 3D geometry construction, object placement, photo assignment, texture generation, and texture propagation. In this paper, we use simple solutions for earlier steps (blue), and focus primarily on the last two steps (orange).}
    \label{fig:task-overview}
\end{figure*}

The \task task involves several steps (see \Cref{fig:task-overview}).
Here, we provide an overview of these steps.

\mypara{Overview.}
Floorplans are usually available as raster images, requiring vectorization.
Raster-to-vector floorplan conversion is the focus of prior work~\cite{Liu2017RastertoVectorRF}, so we assume a vector floorplan as input.
We use `floorplan' to mean a vector floorplan from here on.
We convert the floorplan to 3D geometry and place fixed 3D objects (e.g., doors, windows, toilets) in the floorplan using a rule-based approach that retrieves objects from ShapeNet~\cite{chang2015shapenet} (see supplement for details).
In this work, we focus on computing textures for architectural surfaces in each room, including both surfaces observed in photos and entirely unobserved surfaces.

\mypara{Input and output assumptions.}
The input is a floorplan and a set of photos taken inside a residence, with photos assigned to rooms.
The set of photos are taken from a subset of the rooms.
Some rooms do not have any photos, and some surfaces in a room may not be visible in a photo.
Room type information and correspondence between photos and rooms is commonly available for photos on real-estate websites, so we assume this information in our input.
The floorplan specifies walls and openings (i.e., doors and windows) as line segments, provides a category for each room (e.g., \texttt{bedroom}), and contains structurally fixed objects (e.g., toilets), each of which is given a position.
The output is a 3D mesh of the house with textures for all architectural surfaces in each room.
As a simplifying assumption, we only represent three surface types in each room: `floor', `wall', and `ceiling', indicated by $s \in [1,3]$ and assign the same texture for all surfaces of the same type in a room.
Thus, given the set of rooms of a house $R = \{1,2,\dots, r, \dots, |R|\}$ where $r$ is a room index, we uniquely identify a surface through $(r, s)$.
Each texture is an RGB image assigned to surface $s$ of room $r$, so the complete output texture set is $Y_{r,s} \in \mathbb{R}^{3\times H\times W}$.

A good texture set is \emph{tileable} (i.e. does not exhibit seams or look unnecessarily repetitive when tiled), matches the \emph{color}, \emph{pattern}, and \emph{substance} of the input photo surfaces, while correcting artifacts due to illumination conditions and imaging noise in the input.

%% file: sec/data.tex
\section{Data}

In this section we describe the datasets used in our experiments: a dataset of floorplans and photos based on Rent3D~\cite{ApartmentsCVPR15} that we call Rent3D++, and two curated datasets of textures that we respectively use for training our texture synthesis approach and for establishing a texture retrieval baseline.

\begin{figure}
    \centering
    \includegraphics[width=\linewidth]{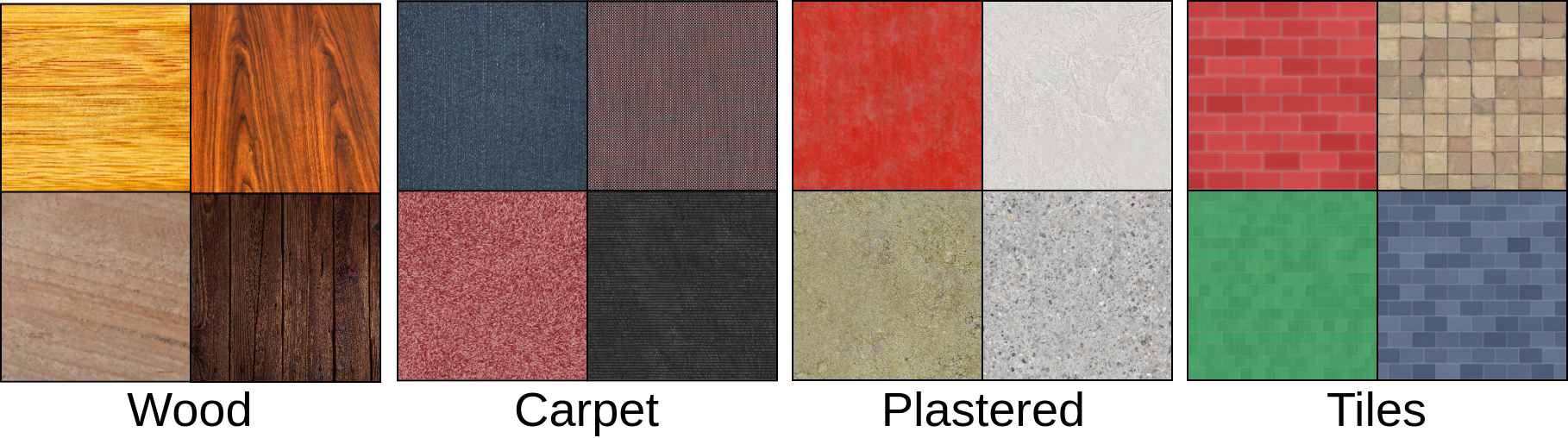}
    \caption{Selected textures from curated textures dataset.}
    \label{fig:data:selected_crops}
\end{figure}

\mypara{Rent3D++ floorplan and photos dataset.}
The Rent3D dataset consists of floorplans and photos from $215$ apartments.
However, we found that: i) some rooms were unannotated; ii) not all portals (windows, doors, room-room openings) are annotated and; iii) some photos were not assigned to rooms.
We correct these issues through re-annotation, and extend the dataset by:
\begin{itemize}\denselist
    \item Fixing incorrectly categorized rooms and adding wall outlines and categories from missing rooms.
    \item Expanding the room category set \texttt{\{reception, bedroom, kitchen, bathroom, outdoor\}} by adding another $7$ common room types: \texttt{\{closet, entrance, corridor, staircase, balcony, terrace, unknown\}}.
    \item Annotating all windows, doors and other wall openings, and associating them with corresponding rooms.
    \item Define a $60/20/20\%$ ($129/43/43$ houses) training, validation, test split (cf. original $100/30/85$ house split) given more samples to training and validation.
    \item Extract rectified surface crops from architectural surfaces seen in photos (floors, walls, ceilings).
\end{itemize}

\begin{figure}
    \includegraphics[width=\linewidth]{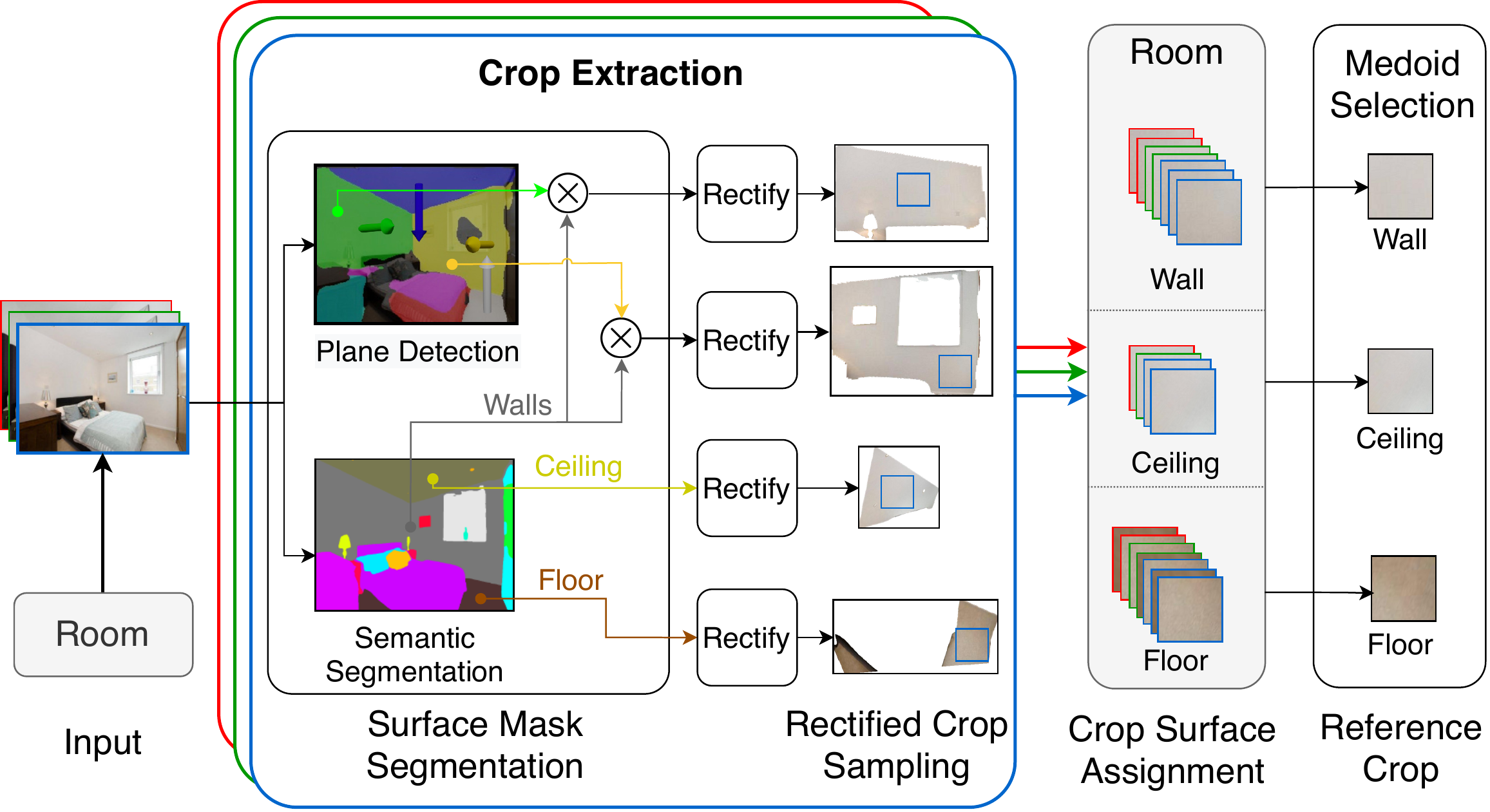}
    \caption{Surface crop extraction approach. We extract rectified surface patches from photos for conditioning texture generation and for use as a reference in our evaluation.}
    \label{fig:data:rectified-crop-sampling}
\end{figure}

\mypara{Surface crop extraction.}
To facilitate texture generation, we extract a set of rectified surface crops for all architectural surfaces observed in photos.
These are square patches which we use to condition the generation of textures and also as `reference crops' for evaluation.
\Cref{fig:data:rectified-crop-sampling} shows the approach we adopt.
We first segment floor, ceiling, and wall surfaces using a semantic segmentation model (HRNet-v2~\cite{hrnet} trained on ADE20K~\cite{ade20k}).
We then use the approach of \citet{yu2019singleimage} to estimate normals and depth for the surface planes ($10$ largest wall masks, one mask for floor and one for ceiling) and rectify the surface masks using the rectification implementation by \citet{bell13opensurfaces} and a predefined constant camera field of view.
Rectified surfaces are upscaled by a factor of $3$ to sample up to $10$ random $256\times256$ crops (max $1000$ attempts to obtain complete crop, per crop).
The crops are resized to $128\times128$ and assigned to the corresponding room surface.
We treat the medoid crop\footnote{By projecting the crops into the $3 \times H \times W$ RGB vector space and selecting the crop closest to the mean vector.} from each surface as a reference crop for evaluation purposes.
The medoid is less likely to be affected by shadows, reflections and specular highlights, while also being representative of the surface.

\mypara{Stationary textures dataset.}
We curated $516$ texture exemplars for four substance types: `wood', `plaster', `carpet' and `tile' from various online texture libraries\footnote{\url{https://www.pexels.com/}, \url{ https://www.sketchuptextureclub.com/}, \url{https://www.freepik.com/}, \url{ https://3djungle.net/}}.
\Cref{fig:data:selected_crops} shows examples from each texture substance type.
We divide this dataset into a training and validation split of $452$ and $64$ textures respectively.
We train our texture synthesis model using crops extracted from these textures so that we can generate stationary textures that can be seam-corrected to tile without artifacts.

\mypara{Substance-mapped textures dataset.}
We also curated a broader dataset of tileable (seamless and stationary) textures from ArchiveTextures\footnote{\url{https://archivetextures.net/}} for our retrieval-based texturing baseline.
It consists of 146 textures from substances such as `carpet', `concrete', `granite', `metal', `painted', `plastic', `tiles' and `wood'. See the supplement for examples.
These textures are seamless and scaled so that they can be directly tiled as textures on our 3D geometry.
Images in this dataset roughly correspond in size to a cropped patch from the stationary textures dataset.

%% file: sec/method.tex
\section{Approach}

Here, we focus on the texture generation and propagation stages in \task.
See the supplement for implementation details of the other stages.

\begin{figure}
    \centering
    \includegraphics[width=\linewidth]{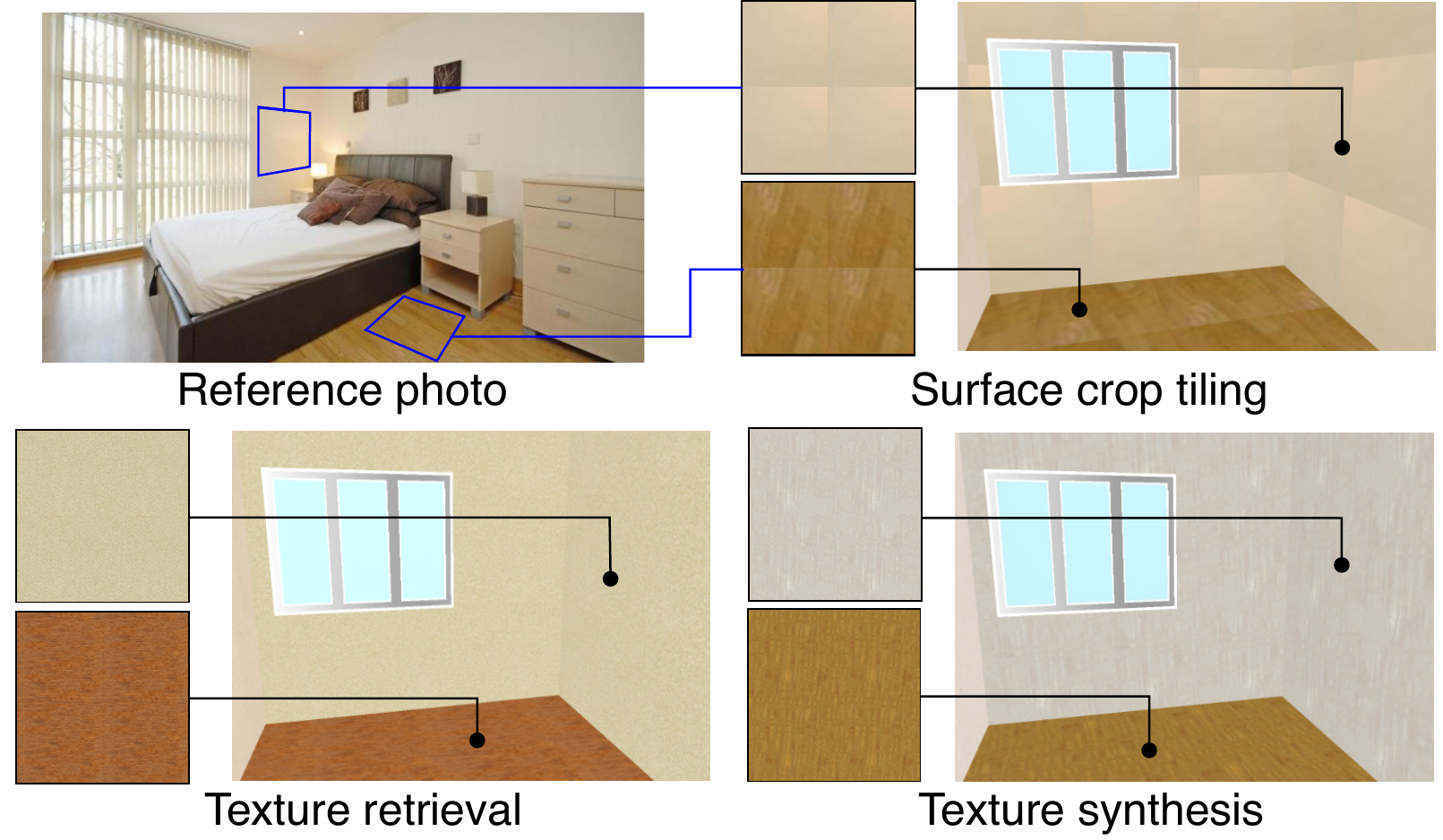}
    \caption{Three approaches to creating textures: i) direct surface crop tiling, ii) retrieval of best matching texture, and iii) texture synthesis conditioned on surface crop. Directly using the crop creates obvious tiling artifacts. Texture retrieval cannot precisely match the surface. Texture synthesis can better match the surface with fewer visible artifacts.}
    \label{fig:method:different_approaches}
\end{figure}

We consider three families of approaches for texture generation: i) direct crop; ii) retrieval; and iii) synthesis.
The first and simplest approach is to directly use the extracted rectified crop from the input surface.
Unfortunately, this can result in obvious seams and repeating artifacts when tiling the texture.
The second approach retrieves a texture that best represents the surface from a dataset.
It can attain high quality output but is limited by the size and diversity of the texture dataset, and may not match the input surface well.
The third approach generates textures conditioned on the input surface and can potentially achieve high quality output that also matches the input surface.
\Cref{fig:method:different_approaches} illustrates these three approaches.

\subsection{Overview}

There are three components to our approach: neural embedding-based tileable texture synthesis, texture synthesis for observed surfaces, and texture propagation to unobserved surfaces.
As the basis to our texture synthesis approach, we enhance an embedding based texture synthesis model~\cite{henzler2020neuraltexture} so that given an input crop, we can embed it to a latent code and synthesize a tileable texture (of finite resolution) from the latent code (\Cref{sec:method:texture-synthesis}).
Using our tileable neural texture synthesis network, we can then take rectified surface crops for observed surfaces and synthesize an appropriate texture (\Cref{sec:method:observed-surfaces}).
As the photos may not cover all surfaces (roughly $60\%$ are unobserved), we create a graph using the room as nodes, and connectivity between the rooms as edges.  We encode the room type and the texture embeddings of the $3$ surface types (`wall', `floor', `ceiling') for the room, and use a graph neural network (GNN) to propagate embeddings to unobserved surfaces.  We can then synthesize texture for all unobserved surfaces using these propagated texture embeddings (\Cref{sec:method:prop}).

\subsection{Learning an embedding for tileable textures}
\label{sec:method:texture-synthesis}

We extend recent work on neural textures synthesis~\cite{henzler2020neuraltexture} to address our problem setting.
To condition our texture on the input surface, we compute a texture embedding $\vec{t} = E(I)$ for each surface crop $I$ using an encoder $E$.
Then, given an embedding $\vec{t}$, we decode a texture $Y = D(\vec{t})$ for each surface.
As this output is not guaranteed to be seamless, we post-process the texture to make it seamless.

\citet{henzler2020neuraltexture}'s original approach accepts a $128 \times 128$ input, which it encodes to an $8$-dimensional embedding.
This embedding is combined with a random noise vector to sample an infinite 2D or 3D texture.
A weakness of this approach is that a separate model is trained for each substance.
Moreover, the approach has difficulty disentangling color, pattern, and substance, especially for surfaces with a variety of colors (e.g., tiles, and painted walls which are common in our setting).

\begin{figure}
    \includegraphics[width=\linewidth]{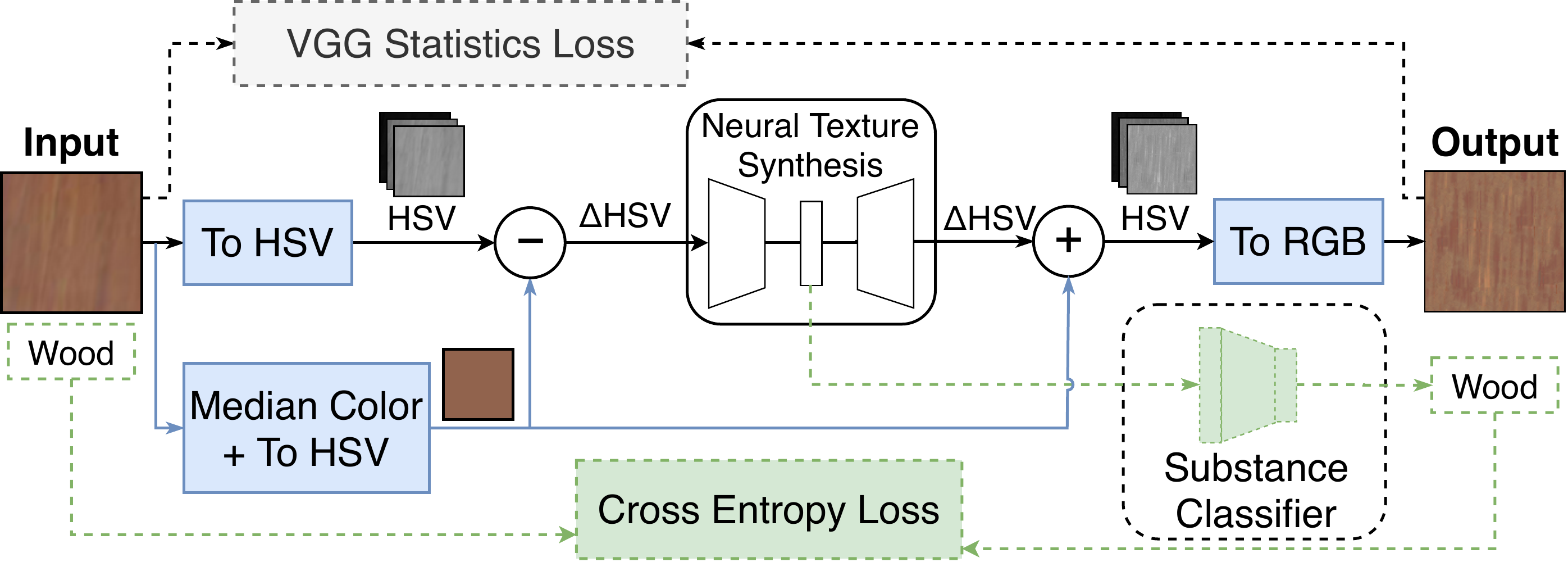}
    \caption{Our texture synthesis architecture employs an encoder-decoder network inspired by recent work~\cite{henzler2020neuraltexture}, adding modules to compute $\Delta$ HSV (blue) and a substance classifier (green). Components that are only used during training have dashed outlines. 
    }
    \label{fig:texture-gen-architecture}
\end{figure}

We extend this approach in several ways to address these limitations (see \Cref{fig:texture-gen-architecture}).
To separate color from texture pattern, we separate into median color and offset and convert both to the HSV color space to obtain the median HSV color and offset from the median $\Delta$HSV.
A differentiable HSV to RGB conversion layer allows gradient back propagation at training time.
To allow the use of a single model across different substances, we also introduce a substance classification branch.
The substance classification branch is trained using a cross-entropy loss over the substance category (`wood', `plaster', `carpet', and `tile'), encouraging a structured latent space based on substance type which is used for texture propagation.
This setup is used to train an embedding for $\Delta$HSV using the sum of the VGG statistics loss from \citet{henzler2020neuraltexture} and the cross-entropy loss above.
We convert the median color to RGB and concatenate it with the learned embedding into a final \emph{texture embedding} $\vec{t} \in \mathbb{R}^{8+3}$ which we use for synthesis of observed surfaces and propagation for unobserved surfaces.

To ensure that the generated texture has stationary statistics (e.g. there are no sharp gradient changes across the image), we train on crops from our stationary textures dataset.
We use random crops from the textures to match inference time when we will be taking random crops from the input surfaces.
We do not train directly on the crops from Rent3D as that produces outputs that are unsuitable for use as tileable textures.
In early experiments, we observed training directly on surface crops tends to produce non-stationary textures with `blob' artifacts caused by light gradients, shadows and specular highlights.
Finally, we post-process the generated texture with the Embark Studios Texture Synthesis Library\footnote{\url{https://github.com/EmbarkStudios/texture-synthesis}} to ensure that they are seamless and exhibit no seams when tiled (see supplement for example).

\subsection{Texture synthesis for observed surfaces}
\label{sec:method:observed-surfaces}
Now that we can embed a single image crop $I$ to a vector $\vec{t} = E(I)$, let us consider how we can synthesize textures for observed surfaces.  Note that for a given surface, we may have multiple crops $\{I_i\}$. From these crops $\{I_i\}$, we compute a single representative \emph{surface texture embedding} $\vec{t}^*$ for each surface.

\begin{figure}
    \centering
    \includegraphics[width=1\linewidth]{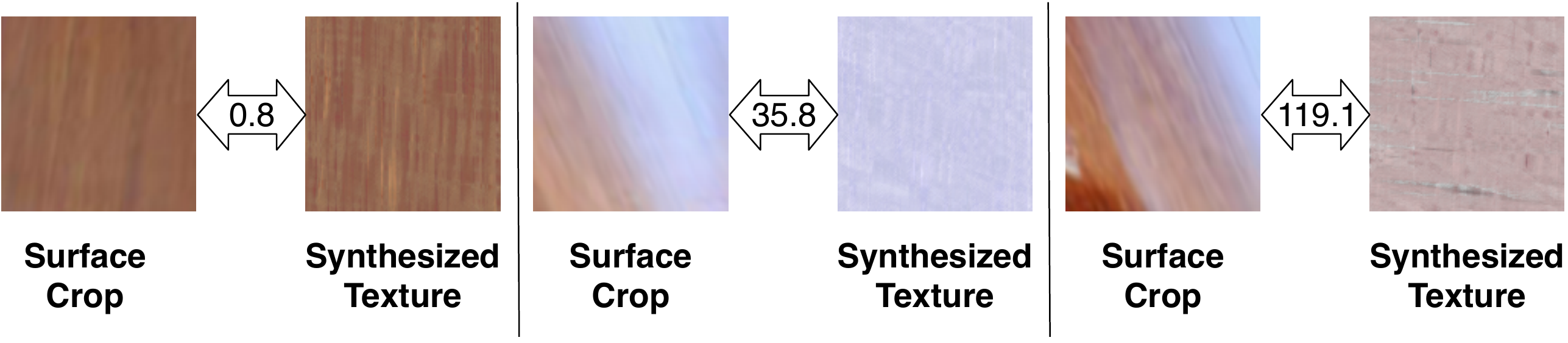}
    \caption{We employ a VGG statistics-based textureness score to characterize the appropriateness of a surface crop as conditioning input for texture generation. The textureness scores are indicated on the arrows (lower is better).}
    \label{fig:textureness}
\end{figure}

A naive approach is to use the mean crop embedding of all the crops assigned to a surface.
This works poorly if a large fraction of crops are affected by artifacts such as shadows and reflections, or if the mean crop interpolates over qualitatively different texture regions.
Furthermore, we want a crop which when embedded will give a good synthesized texture.
Thus, instead of taking the mean, we can select a representative crop for the surface using a textureness score~\cite{Dai2014TheSO,Wu2018AutomaticTE}.

We use the difference of VGG statistics (employed as a similarity metric by \citet{henzler2020neuraltexture}) between the synthesized texture and the conditioned surface crop as a proxy for the textureness score.
For each crop $I_i$, we generate an output texture $D(\vec{t_i})$ and then select the surface crop $I_k$ that has the least L2 difference of VGG Gram Matrices with the synthesized texture.
This textureness score encourages the selection of a surface crop that can be best represented by our texture synthesis model (see \Cref{fig:textureness}).

\begin{figure}
    \includegraphics[width=\linewidth]{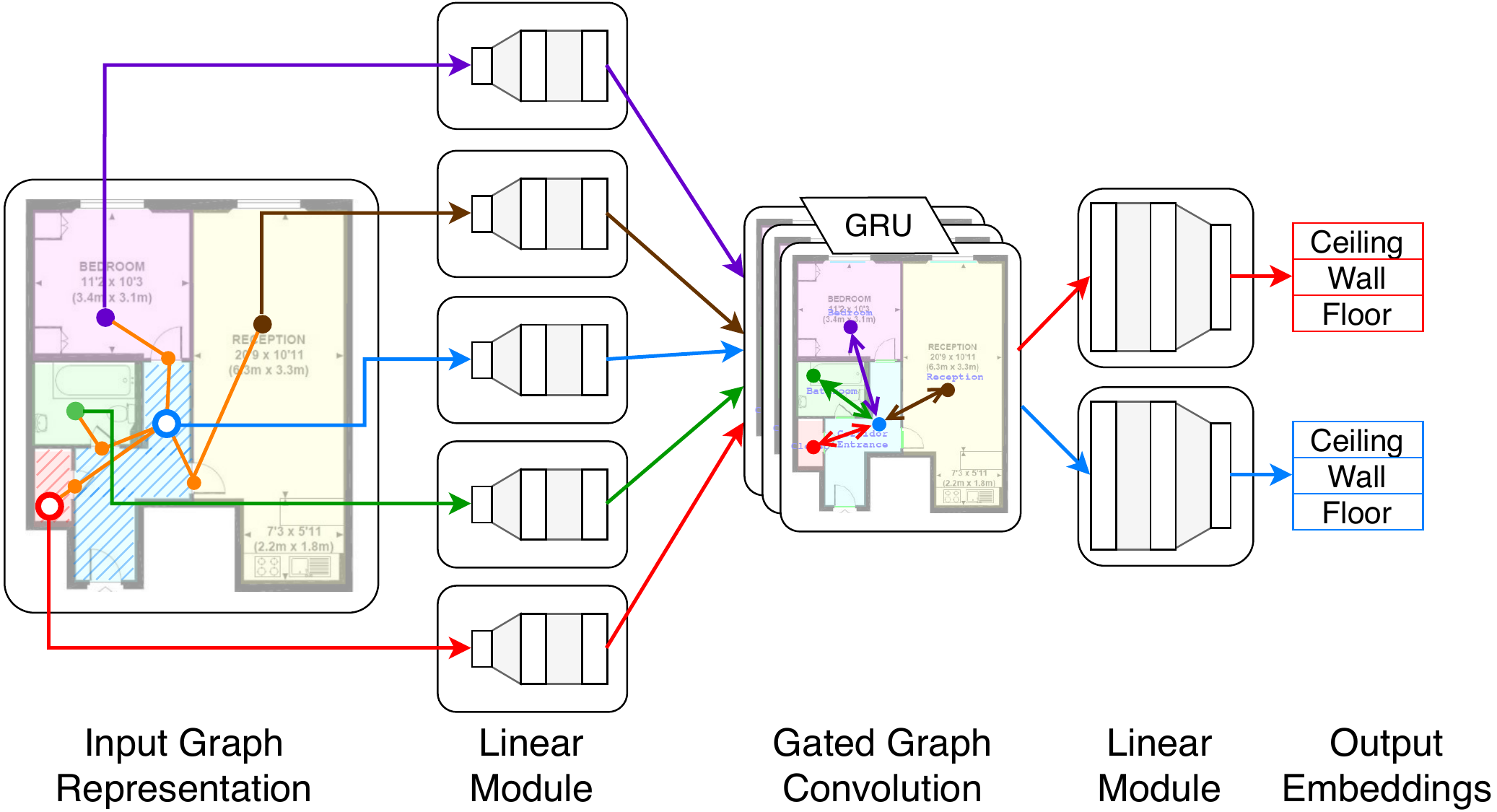}
    \caption{Our graph representation and GNN architecture for texture propagation. Nodes correspond to rooms (with matching colors). Edges indicated by orange lines with kinks at doors. Here, the GNN computes the three surface texture embeddings for the blue node and the red node.}
    \label{fig:method:gnn-architecture}
\end{figure}

\subsection{Texture propagation for unobserved surfaces}
\label{sec:method:prop}

Approximately $60\%$ of the surfaces in our dataset are unobserved by any photo.
For these surfaces, we cannot directly compute $\vec{t}^*$ as above since we do not have any surface crops.
We instead use other observed surface embeddings and the floorplan data (room type and room-door-room connectivity) to propagate a texture embedding.
To do this, we build a room-door-room connectivity graph with rooms as nodes, and edges between two rooms if they are connected by a door.
We encode the room type and the texture embeddings of the three surface types (`wall', `floor', `ceiling') for the room, and use a graph neural network (GNN) to infer the texture embeddings of all unobserved surfaces.

We construct a room-door-room connectivity graph $G=(V,E)$ for each floorplan.
Each room $r$ is represented by a node $v_r$.
An edge exists between rooms connected by a door (see \Cref{fig:method:gnn-architecture}).
Each node $v_r$ stores a feature vector $x_r \in \mathbb{R}^d$ which concatenates the room type $\lambda_r$ as a multi-hot encoding ($\gamma$-dims) and the floor, wall and ceiling surface texture embeddings $\vec{t}^*_{r,\cdot}$ each concatenated with a `presence cell' (indicating if the corresponding surface is observed by setting the cell to $1$, otherwise zeroing out both the embedding and the cell.)
The overall feature vector is $d$-dimensional where $d=\gamma + 3 \times (11 + 1)$.

We use a gated graph convolutional architecture based on \citet{Li2016GatedGS}, with additional linear layers (see \Cref{fig:method:gnn-architecture}).
To train this GNN, we take each observed surface $s$ of each room $r$ in the training set as unobserved and treat it as a prediction target.
As data augmentation, we mask additional surfaces to produce versions of the graph with additional unobserved surfaces.
We use the L1 loss on the target surface texture embedding $\vec{t}^*_{r,s}$ and train with Adam~\cite{kingma2014adam} (weight decay $0.0001$, batch size $32$, learning rate $0.0005$).
The architecture was implemented using PyTorch Geometric~\cite{fey2019fast}.
See the supplement for the implementation details.

%% file: sec/experiments.tex
\section{Experiments}

\begin{figure*}
    \centering
    \includegraphics[width=\textwidth]{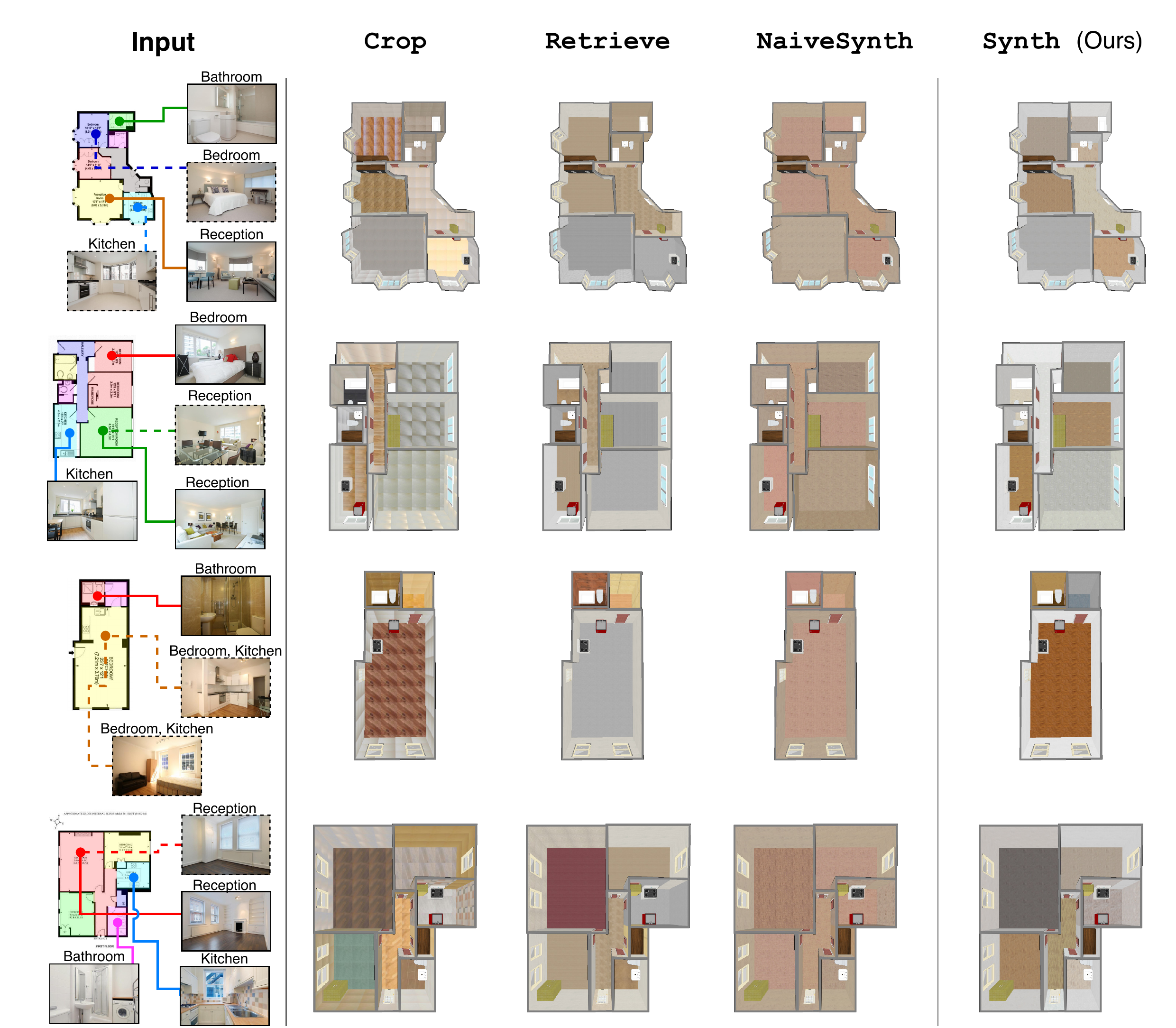}
    \caption{Qualitative comparison of results on test set, with simulated unobserved photos (dashed lines indicate unobserved photos). \Crop produces textures that do not tile well. \Retrieve is a competitive baseline, but does not match the input as well as our \Synth approach does.}
    \label{fig:eval:qualititative-multiple}%
\end{figure*}

\input{results_tables/overall_baselines}

\subsection{Methods}
We compare the following methods in our experiments.

\mypara{\Crop}:
Assign surface reference crop as texture. For unobserved surfaces, we randomly select a crop with the same room and surface type (from the house if available, else from training set, relaxing room type match if necessary).

\mypara{\Retrieve}:
Retrieve textures in the Substance Mapped Textures dataset with lowest pixel-wise L1 loss to the reference crop.
For unobserved surfaces, we employ a similar input selection strategy as \Crop.

\mypara{\NaiveSynth}:
Synthesize textures using \citet{henzler2020neuraltexture} approach trained on our textures dataset. For observed surfaces, conditioning input is the mean crop embedding. For unobserved surfaces, we compute a mean crop embedding considering surface crops assigned to all the surfaces in the training set, of the same room type and surface type. Finally, we correct the seams of synthesized texture crops as before to make them tileable.

\mypara{\Synth}:
Our improved texture synthesis module for observed surfaces leveraging the textureness score, using propagation of improved embeddings with our GNN architecture to handle unobserved surfaces.

\subsection{Metrics}

Texture synthesis approaches are typically evaluated in terms of similarity to reference images and output diversity.
In our setting, we want to `match' the input surface crops while correcting artifacts such as shadows, specular highlights, and making the texture tileable.
These criteria make pixel-wise similarity metrics insufficient, and diversity induced by a random noise vector is inappropriate.
We define a suite of metrics that measure: i) how well a texture matches properties of the input surface such as color, pattern, and substance type; ii) the tileability of the textures; and iii) how well the texture set distribution matches the input surface distribution.
The following metrics define each of these axes of output quality against the reference crop for each surface (or all crops from a surface for \FID).

\mypara{\Color}:
L1 distance between histograms of pixel values in HSL color space ($10$ bins for hue, $3$ bins each for saturation and lightness), normalized to $[0,1]$.

\mypara{\Freq}:
Measures pattern match.
Extract periodic component of grayscale image in frequency domain~\cite{Moisan2010PeriodicPS} and take L1 difference of azimuthally averaged frequency amplitude histograms (with DC component set to zero), taking mean across frequencies.

\mypara{\Subs}:
Substance classification error rate.
Uses a VGG16-based network to classify whether texture matches reference crop's substance type.
See supplement for details.

\mypara{\FID}:
Fréchet Inception Distance~\cite{Heusel2017GANsTB}.
Measures distribution similarity between synthesized texture set and all surface crops.
Lower is better.

\mypara{\Tile}:
Measures uniformity of local color averages as a proxy for tileability.
Based on `mean prior loss'~\cite{Aittala2016ReflectanceModelling}, and implemented as $\text{Tile}(I_\text{gray}) = || w_{f} \odot F\{I_\text{gray}\}||^2$ where $F$ is the Fourier operator, $I_\text{gray}$ is crop $I$ in grayscale, and $w_f$ is the magnitude spectrum of a Gaussian with $\sigma=21$ ($1/6$ texture size, as recommended by \cite{Aittala2016ReflectanceModelling}).
Lower is better.

\subsection{Quantitative Evaluation}

\Cref{tab:overall-results} shows the overall performance of our method compared against various baseline approaches.
Similarity metrics for the `all' and `unobserved' surfaces settings are computed by simulating a further 60\% unobserved photos since we do not have reference crops for truly unobserved surfaces.
\Crop does well on similarity metrics by virtue of directly using the reference crop as a texure.
However, it does not produce tileable textures as seen by the high \Tile metric values.
Overall, our \Synth outperforms other baselines on both observed and unobserved surfaces.
The \Retrieve approach is quite competitive and even outperforms \NaiveSynth on many metrics.

The supplement provides additional analyses and ablations showing that these trends hold across a range of unobserved photo fractions, and quantifying improvements due to individual components in our approach.

\subsection{Qualitative Evaluation}

\Cref{fig:eval:qualititative-multiple} compares full textured 3D house results using various methods.
\Crop produces obvious tiling artifacts such as notable repetitiveness and visible seams.
\NaiveSynth does a poor job in capturing the input appearance, in this case generating red shifted textures.
\Retrieve produces high quality textures, but it fails to match the input as well as our \Synth method.
For instance looking at the green shaded bathroom (which is observed) and the blue shaded bedroom (which is unobserved) in the first example, cyan room (observed) in the second example, yellow room (unobserved) in the third example, and red room (observed) in the last example, \Synth better matches the appearance of the floor.
See the supplement for more examples.

\subsection{User Study}

To holistically evaluate the results of our approach against baselines, we also carry out a user study.

\topic{Setup.}
We randomly sampled 20 houses from the test set and conducted a forced A-B choice user study.
The participants were university students who were not involved with this work.
They were asked to choose between two top-down 3D renderings of a textured house output (with the ceiling removed), given an input floorplan overlaid with photos assigned to the rooms.
The pair of renderings was from \Synth and one of the three baselines, presented in random order.
Users were instructed to consider the quality of the textures (i.e. absence of discontinuities and unnecessary repetitiveness), and similarity to the surfaces observed in the photos.
As the pairs contrasted the three baselines against \Synth, each user made $20\times 3 = 60$ choices in shuffled order.

\mypara{Results.}
A total of $18$ users participated the study.
Our \Synth results were preferred relative to the baselines about $70$\% of the time ($69.4$\% against \Retrieve, $66.9$\% against \Crop and $72.8$\% against \NaiveSynth).
See the supplement for additional analysis of the user study results.

\subsection{Failure cases and limitations}

\begin{figure}
    \centering
    \subfloat[\centering Semantic Mask Failure]{\includegraphics[width=0.5\linewidth]{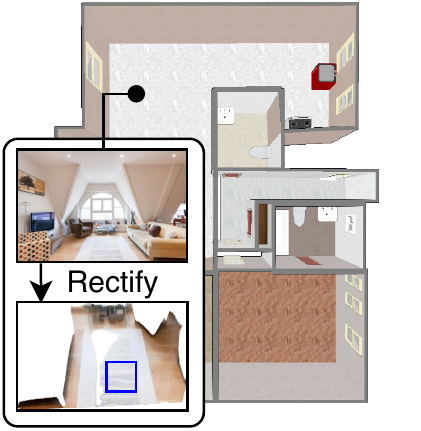}}
    \subfloat[\centering Illumination Failure]{\includegraphics[width=4.28cm]{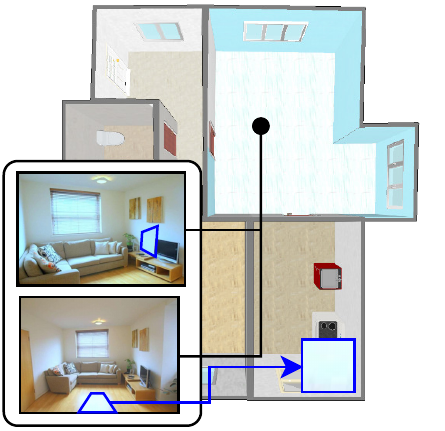}}
    \caption{Example failure cases. Selected crops shown in blue outline. In the first case, a semantic segmentation issue leads to a carpet crop being used for the wooden floor. In the second case, a blue color shift in the photos and strong specular highlights cause poor crop selection.}
    \label{fig:exp:user-study:failures}
\end{figure}

\Cref{fig:exp:user-study:failures} shows two failure cases.
In the first case, a wooden floor is textured entirely as a white carpet.
Here, the semantic segmentation we use included the carpet mat in the floor mask, which led to a crop from the carpet being used for floor texture synthesis.
In the second case, the walls exhibit a blue tint due to the illumination and color balance of the input photo.
Our crop selection helps mitigate localized illumination anomalies, but it does not correct photo-wide color shifts, or otherwise handle the general problem of accounting for illumination.
The floor here is also light cyan due to poor crop selection caused by a severe specular highlight (crop shown in popup is entirely solid color and is likely interpreted as a plastered surface).

%% file: results_tables/overall_baselines.tex
\begin{table*}
\ra{1.3}
\centering
\caption{Overall results on test set. Lower values are better. Similarity metrics in the `Unobserved' and `All' columns are reported by simulating 60\% of photos as unobserved. All similarity metrics treat the reference crop as the ground truth (note that this benefits \Crop which uses the reference crop as a texture). We observe that \Synth outperforms \Retrieve and \NaiveSynth across all metrics, and outperforms all approaches in terms of tileability.
}
\label{tab:overall-results}
\resizebox{\linewidth}{!}{
\begin{tabular}{@{}l rrrrr@{\hspace{7mm}}rrrrr@{\hspace{7mm}}rrrrr @{}}
\toprule
& \multicolumn{5}{c}{Observed} & \multicolumn{5}{c}{Unobserved} & \multicolumn{5}{c}{All} \\
\cmidrule(l{0mm}r{7mm}){2-6} \cmidrule(l{0mm}r{7mm}){7-11} \cmidrule(l{0mm}r{0mm}){12-16}
& \Color & \Freq & \Subs & \FID & \Tile & \Color & \Freq & \Subs & \FID & \Tile & \Color & \Freq & \Subs & \FID & \Tile\\
\midrule
\Crop & 0 & 0 & 0 & 0 & 38.1 & 0.768 & 0.026 & 0.345 & 57.2 & 40.6 & 0.459 & 0.016 & 0.208 & 35.6 & 39.5\\
\midrule
\Retrieve & 0.561 & 0.054 & 0.473 & 238.2 & 17.3 & 0.751 & 0.040 & 0.437 & 261.5 & 19.1 & 0.680 & 0.046 & 0.458 & 243.2 & 18.3\\
\NaiveSynth & 0.694 & 0.046 & 0.385 & 239.3 & 21.7 & 0.714 & 0.044 & 0.427 & 245.4 & 19.8 & 0.709 & 0.046 & 0.404 & 239.4 & 20.6\\
\Synth (ours) & \bf{0.431} & \bf{0.035} & \bf{0.350} & \bf{196.1} & \bf{16.4} & \bf{0.653} & \bf{0.032} & \bf{0.393} & \bf{199.4} & \bf{18.6} & \bf{0.591} & \bf{0.034} & \bf{0.392} & \bf{196.2} & \bf{17.6}\\
\bottomrule
\end{tabular}
}
\end{table*}

%% file: sec/conclusion.tex
\section{Conclusion}

We presented the \task task: conversion of a floorplan and set of sparse photos to a textured 3D mesh.
We focused on the texture generation stages in this task, defining a suite of texture quality metrics, and proposing a texture synthesis approach with GNN-based propagation to unobserved surfaces.
Our experiments show that our approach leads to higher quality textures compared to simpler baselines.
Several stages of the task were simplified and are open for future investigation, including improved generation of regular patterns such as tiles.
We believe conversion of floorplan and photo data to textured 3D scenes is a promising avenue for creating large volumes of 3D interiors, and will enable future work relying on large-scale 3D interior datasets.

%% file: sec/acks.tex
\vspace{1em}
\mypara{Acknowledgements.}
{
We thank all the participants in our user study: Akshit Sharma, Andy Wang, Hanxiao Jiang, Hao Hao, Jiaqi Tan, Julia Read, Kevin Joseph, Leon Kochiev, Liyang Zhou, Mayur Mallya, Reza Asad, Richard Pan, Roya Javadi, Sachini Herath, Saghar Irandoust, Sebastian Dille, Sepideh Sarajian, Sepidehsadat Hosseini, Shivansh Patel, Supriya Gadi Patil, Vishal Batvia, Weijie Lin, Weilian Song, Yasaman Etesam, Yongsen Mao, Yuzhen Mao. This research was enabled in part by support provided by \href{www.westgrid.ca}{WestGrid} and \href{www.computecanada.ca}{Compute Canada}.
QW is supported by a VSP USRA, AXC by a Canada CIFAR AI Chair, MS by a Canada Research Chair and NSERC Discovery Grant.
}